\documentclass[10pt,twocolumn,letterpaper]{article}

\usepackage{times}
\usepackage{epsfig}
\usepackage{graphicx}
\usepackage{amsmath}
\usepackage{amssymb}
\usepackage{xspace}

\newcommand{\etal}{\emph{et al.}\xspace}

\begin{document}

\title{Deep Secure Encoding: An Application to Face Recognition}

\author{Rohit Pandey \and Yingbo Zhou \and Venu Govindaraju
}

\date{}

\maketitle
\thispagestyle{empty}

\begin{abstract}   
   In this paper we present Deep Secure Encoding: a framework for secure classification using deep neural networks, and apply it to the task of biometric template protection for faces. Using deep convolutional neural networks (CNNs), we learn a robust mapping of face classes to high entropy secure codes. These secure codes are then hashed using standard hash functions like SHA-256 to generate secure face templates. The efficacy of the approach is shown on two face databases, namely, CMU-PIE and Extended Yale B, where we achieve state of the art matching performance, along with cancelability and high security with no unrealistic assumptions. Furthermore, the scheme can work in both identification and verification modes.

\end{abstract}

\section{Introduction}
Biometric template protection is an important factor in making the deployment of biometric authentication as widespread as string based password security. Authentication on the basis of ``who we are" instead of ``something we possess" or ``something we remember", offers convenience and often, stronger system security. The advantages of biometric authenticators are straightforward to see but the disadvantages require further thought.

A typical biometric authentication system would use a few samples of user's biometric modality (e.g.\ image of face, fingerprint or iris) for enrollment, and extract and store a template of the user from them. During authentication, verification or identification is performed where a given sample is matched against the stored templates and depending on the matching score, access is granted or denied. In order to motivate the importance of storing the enrolled templates in a secure and cancelable manner, we compare a biometric template to a string based password. 

When registering a string password, a one way non-invertible transform (i.e.\ a hash) of it is stored. During verification, a password is entered and it's hash value is calculated. The hash is compared with the stored hash and if the two strings matched exactly, their hashes would match as well, and access would be granted. In such a scenario, the stored hash reveals no information about the original password and also, if the password is compromised, it can be changed and a new hash can be stored. This is the kind of security we desire from biometric templates as well. Unlike passwords, biometric modalities lack two important aspects. 1) They rarely match exactly between different readings, and 2) they cannot be changed if compromised. Thus, the objective of biometric template protection schemes is to extract authenticators from biometric modalities that are 1) secure i.e.\ given the authenticator, it should be infeasible to extract any information about the original modality, and 2) cancelable i.e.\ if compromised, it should be possible to extract a new authenticator from the same modality.


\section{Previous work}
These objectives have been tackled for faces in different ways. Schemes that used cryptosystem based approaches but without hash functions include Fuzzy commitment schemes by Ao and Li~\cite{ao2009near}, Lu \etal~\cite{lu2009face} and Van Der Veen \etal~\cite{van2006face}, and fuzzy vault by  Wu and Qiu~\cite{wu2010transforming}. The fuzzy commitment schemes suffered from limited error correcting capacity or short keys in general. Fuzzy vault schemes suffer from the problem that the data is stored in the open between chaff points, and also that they cause an overhead in storage space. Some quantization schemes were used by Sutcu~\etal~\cite{sutcu2007protecting,sutcu2005secure} to generate somewhat stable keys. There were also several works that combine the face data with user defined keys or user specific keys. These include combination with a password by Chen and Chandran~\cite{chen2007biometric}, user specific token binding by Ngo~\etal~\cite{ngo2006biometric,teoh2006random,teoh2004personalised}, biometric salting by Savvides~\etal~\cite{savvides2004cancelable}, and user specific random projection schemes by Teoh and Yuang~\cite{teoh2007cancelable} and Kim and Toh~\cite{kim2007method}. Hybrid approaches that combine transform based cancelability with cryptosystem based security like \cite{feng2010hybrid} have also been proposed but give out user specific information to generate the template creating possibilities of masquerade attacks. Pandey and Govindraju \cite{pandey2015secureface} proposed a security centric scheme that used features extracted from local regions of the face to obtain exact matching and thus, benefited from the security of hash functions like SHA-256. Although seemingly more secure, the matching accuracy of the scheme was low and the feature space being hashed was not uniformly distributed.

Here, we propose a scheme that achieves state of the art matching accuracy while maintaining a very high level of security with no unrealistic assumptions. Furthermore, the scheme can work in both verification and identification modes while being truly key-less. The contributions of this paper are as follows. 

1. We design a secure classification scheme that achieves state of the art matching performance on CMU-PIE and Extended Yale B databases, while achieving template cancelability and security.

2. We address the challenges of learning deep neural networks with limited data in a biometric setting, as well as, that of generating a ranged, tunable score from a classification framework.

3. We provide an analysis of the cancelability and security of the generated templates in two scenarios and show that the second, more secure scenario, achieves high security without any assumptions on the data distribution or any need for parameter protection.


\begin{figure*}
	\begin{center}
	\includegraphics[width=0.6\textwidth]{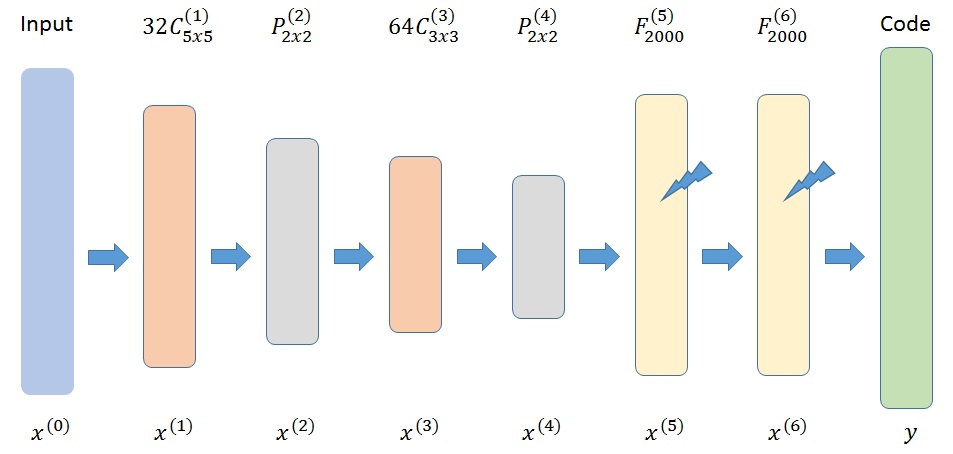}
	\end{center}
	\caption{An illustration of the deep secure encoding network used in this work.  We use blue, orange, gray, yellow and green rectangle to denote input, convolution layer, pooling layer, fully connected layer and the output, respectively.  The lightning shape is used to denote stochasticity in the layer. (Figure best viewed in color)}
	\label{fig:network}
\end{figure*}

\section{Method}
Our work is inspired by the error correcting output codes for multi-class classification, where one partitions a $k$-class problem to multiple binary class problems \cite{dietterich1995solving}.  In addition, this framework relies on the superior feature learning capability of deep convolutional neural networks (CNNs) to learn this mapping from face images to the codes assigned to each user.  We exploit the classification performance of the CNN along with the high entropy of randomly generated binary codes for each class to design an architecture with high matching accuracy as well as high template security, with minimal or no assumptions. We proceed to describe the method in detail.

\subsection{Convolutional neural network}
Convolutional neural networks (CNN) \cite{Lecun98gradient-basedlearning} are biologically inspired models, which contain three basic components: convolution, pooling and fully connect layers.  In the convolution layer one tries to learn a filter bank given input feature maps.  The input of a convolution layer is a 3D array with $d$ number of 2D feature maps of size $n_1\times n_2$.  Let $x_{ijk}$ denote the component at row $j$ and column $k$ in the $i$th feature map, and we use $x_i^{(l)}$ to denote the complete $i$th feature map at layer $l$.  If one want to learn $h_f$ set of filters of size $f_1 \times f_2$, the output $x^{(l+1)}$ for the next layer will still be a 3D array with $h_f$ number of 2D feature maps of size $(n_1-f_1+1) \times (n_2-f2+1)$.  More formally, the convolution layer computes the following:
\begin{equation}
x_j^{(l+1)} = s(\sum_i F_{ij} * x_i^{(l)} + b_j)
\end{equation}
where $F_{ij}$ denotes the filter that connect feature map $x_i$ to output map $x_j^{(l)}$, $b_j$ is the bias for the $j$th output feature map, $s(\cdot)$ is some elementwise non-linearity function and $*$ denotes the discrete 2D convolution.  We denote $k$-th convolutional layer with $h_f$ filters of size $f_1\times f_2$ by $h_fC^{(k)}_{f_1\times f_2}$

The pooling (or subsample) layer takes a 3D feature map and tries to down-sample/summarize the content with less spatial resolution.  Pooling is done for every feature map independently and with non-overlapping windows.  An intuition of such operation is to have some build in invariance against small translation as well as reduce the computation for upper layers.  For example, with a pooling size of $p_1 \times p_2$ an input with $h_f$ number of 2D feature maps of size $n_1 \times n_2$ will result an output of $h_f$ number of feature maps but of size $n_1/p_1 \times n_2/p_2$.  For average (mean) pooling, the output will be the average value inside the pooling window, and for max pooling the output will be the maximum value inside the pooling window.  We denote the $k$-th pooling layer with pooling window size $p_1 \times p_2$ by $P^{(k)}_{p_1\times p_2}$.

The fully connected layer connects all the input units from the lower layer $l$ to all the output units in the next layer $l+1$.  In more detail, the next layer output is calculated by:
\begin{equation}
x^{(l+1)} = s(Wx^{(l)} + b)
\end{equation}
where $x^{(l)}$ is the vectorized input from layer $l$, $W$ and $b$ are the parameters of the fully connected layers.  We denote the $k$-th fully connected layer with $h$ number of hidden units by $F^{(k)}_h$.

A CNN is commonly composed by several stacks of convolution and pooling layers followed by a few fully connected layers.  During training the last layer is associated with some loss functional to provide training signals, and the training for CNN can be done by using gradient descent.  For example, in classification the last layer is normally a softmax layer and cross entropy is commonly used as the loss function during training. 

\subsection{Deep secure encoding}
For secure encoding, it is desirable to obtain a code that is of high entropy so that the code is unlikely to reveal any information regarding the encoded information.  The maximum entropy of a $k$-bit binary code will be obtained if we generate the code with 0.5 probability of each bit taking value one.  A natural solution for secure encoding is therefore to generate different binary code randomly for each of the class and then try to learn the mapping from the data to its corresponding code.  This idea is very similar to the idea in error correcting output codes \cite{dietterich1995solving} if we view each bit as a binary partitioning of the dataset.

We take advantage of the superior learning ability of the neural networks and try to learn this mapping from data directly.  Since the code is binary, for the last layer of the neural network we apply the logistic function ($f(x)=(1+\exp^{-x})^{-1}$) to get values between zero and one.  Cross entropy is used as the loss during training.

To obtain better generalization error, we employed stochastic layers at higher layers of the network.  The stochastic activation function that we employed has the following form:
\[
\tilde{s}(x,\eta) = \eta + s(x+\eta)
\]
where $\eta$ is a independent noise variable from some distribution, and $s(\cdot)$ is some elementwise non-linear function.
The stochastic layer is able to model more complicated distributions including mutimodal distributions, which is desired here.  In addition, it also has the potential benefit of providing better generalization performance, since adding noise has been found to be a good way of regularization \cite{SietsmaD91}.  With the stochastic layer, this network can also be viewed as an ensemble of binary classifier networks that share parameters.  Similar as presented in paper \cite{Bengio14GSN}, here we use the method called ``straight-through", which estimates the gradient of a stochastic unit by just considering its differentiable parts.  The architecture that we used in this work is as follows: two convolutional layers of $32$ filters of size $5\times 5$ and $64$ filters of size $3\times 3$, each followed by max pooling layers of size $2\times 2$. The convolutional and pooling layers are followed by two fully connected stochastic layers of size $2000$ each, and finally the output.  We use activation function $s(x) = \max (x, 0)$ for all layers, and for the stochastic layers we set $\eta \sim \mathcal{N}(0, 2)$.  The network structure is illustrated in Figure \ref{fig:network}.



\subsubsection{Data augmentation}
In a biometric setting, we usually have limited enrollment samples and thus, augmentation of the training set is required for good performance from the model.  For each training example of size $m\times m$, we perform the following augmentation. Instead of using images of the original size, we use partial crops of the original image, which also provide enough information for classification purposes. All possible crops of size $n\times n$ are extracted from the original image yielding $(m-n+1)\times(m-n+1)$ crops. In addition, we flip each crop along the vertical axis to obtain an additional $(m-n+1)\times(m-n+1)$ images. Thus, each image is augmented to $2\times(m-n+1)\times(m-n+1)$ images in order to give us more examples to train the CNN.

\subsection{Secure template generation}
Now that we have a trained neural network that learns how to map face classes to our secure codes $y$, we can proceed to generating secure templates from the codes. We consider two scenarios for this purpose. The first uses the output of the network as it is, and the second quantizes the output and hashes it using SHA-256 to generate a more secure template. From here on we refer to these two scenarios as scenario 1 and 2.

\subsubsection{Scenario 1}
In scenario 1, we use the output of the neural network as it is, without any hashing. Thus, during enrollment, we save the codes assigned to each class as our template. During verification, a sample is fed through the network and the output is now a real valued vector with values between 0 and 1. This is compared to the stored code for the class and some distance measure is calculated against it yielding a real valued score. The sample may now be accepted or rejected depending on the threshold selected for the score. Similarly for identification, the score can be calculated against each of the stored codes and the class yielding the best score can be identified as the face class. In our case this score is proportional to the negative of the euclidean distance between the network output and the stored code. The negative, or more specifically $max(scores) - score$, is taken as the score in order to maintain a protocol of the higher score being better. Next, instead of doing this once for a face sample, we take several crops of the sample (similar to our training process) and calculate the average score for all crops. Thus, for a test sample of size $m\times m$ we use all possible crops of size $n\times n$ along with their corresponding flipped crops. Now our score will be the mean of the scores of obtained for all $2\times (m-n)\times (m-n)$ crops. Note that for security purposes, this scenario best works in situations where the network parameters can be protected. We analyze this aspect in further detail in the security analysis section.

\subsubsection{Scenario 2}
For scenario 2, we quantize and hash the output of the neural network to obtain high security. During enrollment, instead of storing the codes for each class, we store the SHA-256 hash of the codes. While verification, a sample is fed through the network yielding a real valued vector with values between 0 and 1. This output is now quantized using $0.5$ as a threshold to yield a binary vector. If our network has been trained properly, and the given sample is sufficiently close to the claimed class, this code should correspond to the code assigned to the class during training. Next, we take the SHA-256 hash of the quantized output and compare it to the stored SHA-256 hash of the code assigned to the class. In order to yield a score that can take a range of values instead of just a binary yes or no, we use a technique similar to that of scenario 1. Instead of using the full sized image, we use all possible crops of the sample and count the number of crops whose hashed outputs match correctly. This yields a discrete score in the range of $0$ to $2\times (m-n)\times (m-n)$. Doing so makes the system score tunable and allows us to set the security level of the system. In this scenario the network parameters need not be protected and we do not need to make any assumptions to guarantee the security of the system. This is explained in detail in the security analysis section.


\section{Experiments}

We now proceed to give an overview of the datasets used for our experiments and explain the evaluation protocols used, as well as, the specifics of the parameters used for experimentation.

\begin{figure}
	\begin{center}
		\includegraphics[width=\linewidth]{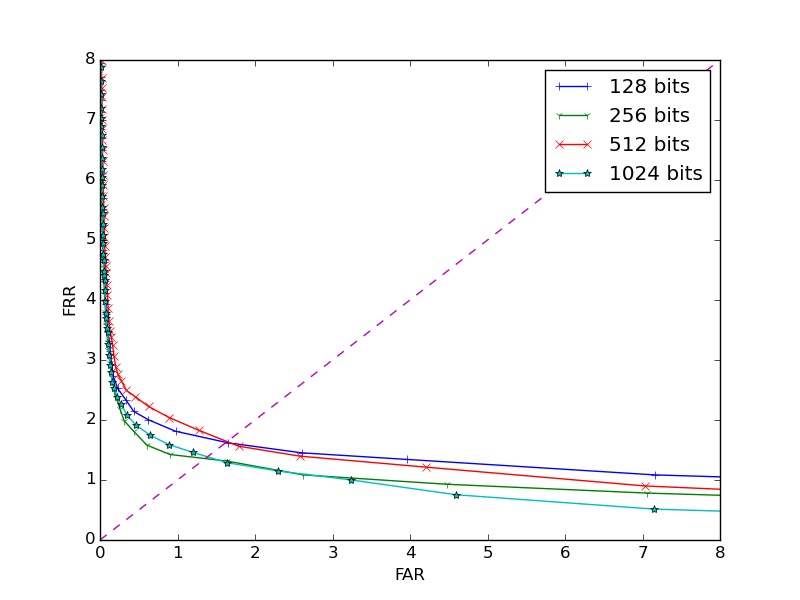}
	\end{center}
	\caption{ROC curves for CMU-PIE (scenario 1) with varying bits of security.}
	\label{fig:rocpie1}
\end{figure}

\begin{figure}
	\begin{center}
		\includegraphics[width=\linewidth]{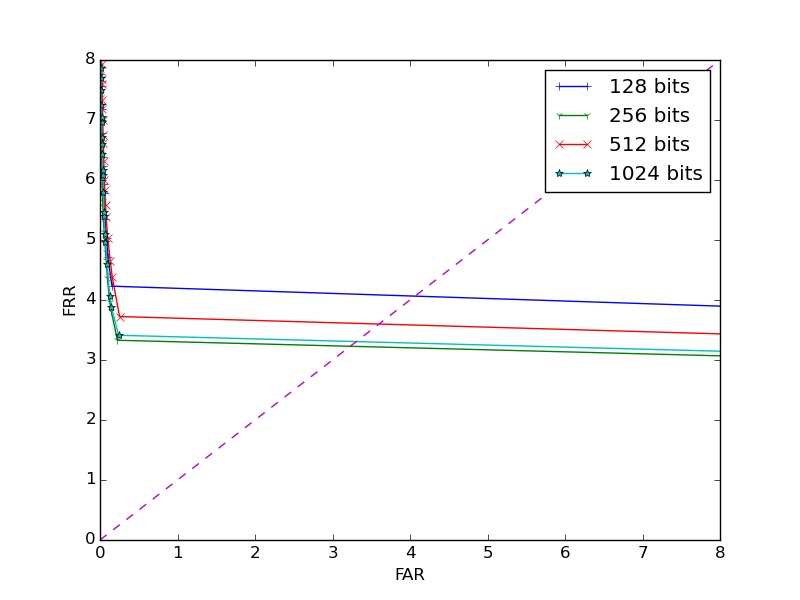}
	\end{center}
	\caption{ROC curves for CMU-PIE (scenario 2) with varying bits of security.}
	\label{fig:rocpie2}
\end{figure}

\begin{figure}
	\begin{center}
		\includegraphics[width=\linewidth]{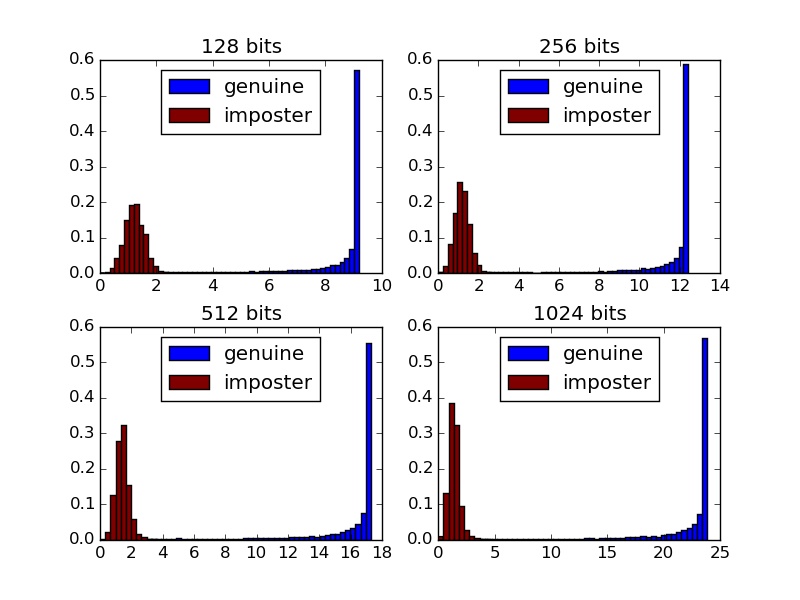}
	\end{center}
	\caption{Genuine and imposter distributions for CMU-PIE (scenario 1) with varying bits of security.}
	\label{fig:genimppie1}
\end{figure}

\begin{figure}
	\begin{center}
		\includegraphics[width=\linewidth]{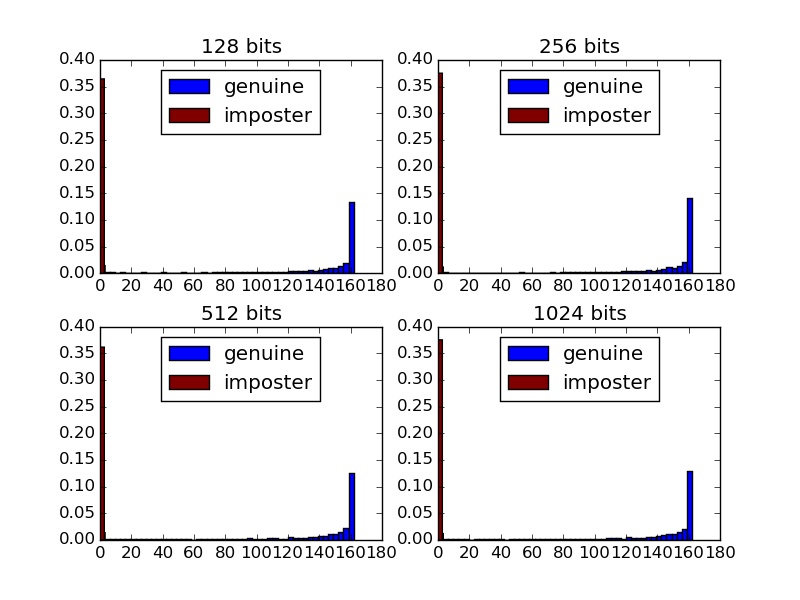}
	\end{center}
	\caption{Genuine and imposter distributions for CMU-PIE (scenario 2) with varying bits of security.}
	\label{fig:genimppie2}
\end{figure}

\begin{table*}[t]
	\begin{center}
	\begin{tabular}{|c|c|c|c|c|}
		\hline
		Bits of security & EER (scenario 1) & Accuracy (scenario 1) & EER (scenario 2) & Accuracy (scenario 2) \\ \hline \hline
		128 & 1.61\%                              & 95.76\%                                     & 4.05\%                              & 95.60\%                                     \\ 
		256 &  1.36\%                             & 95.93\%                                     & 3.23\%                              &  95.68\%                                     \\ 
		512 & 1.63\%                              & 95.23\%                                     & 3.59\%                              & 95.11\%                                     \\ 
		1024 & 1.39\%                               & 95.82\%                                     & 3.31\%                              & 95.67\%                                     \\ \hline
	\end{tabular}
	\end{center}
	\caption{Identification accuracy and equal error rate on CMU-PIE database with varying bits of security.}
	\label{tab:pie}
\end{table*}

\begin{figure}
	\begin{center}
		\includegraphics[width=\linewidth]{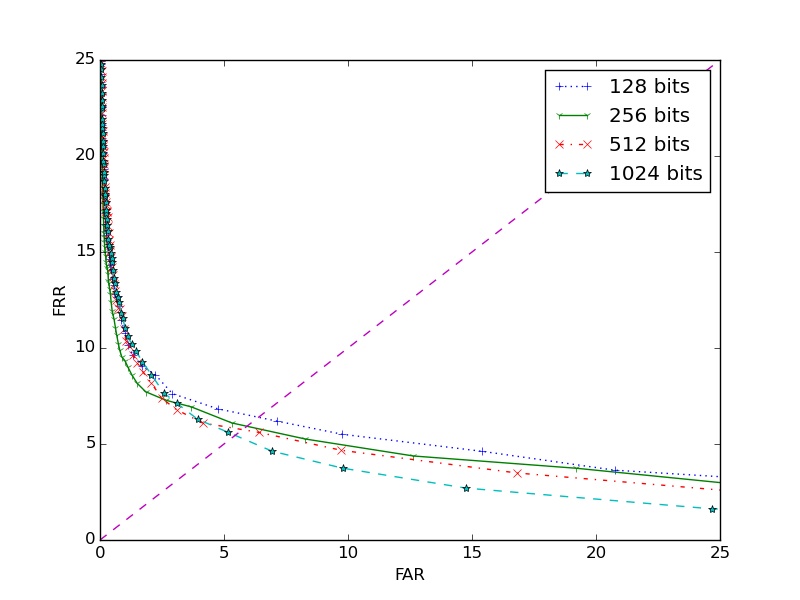}
	\end{center}
	\caption{ROC curves for Extended Yale B (scenario 1) with varying bits of security.}
	\label{fig:yaleroc1}
\end{figure}

\begin{figure}
	\begin{center}
		\includegraphics[width=\linewidth]{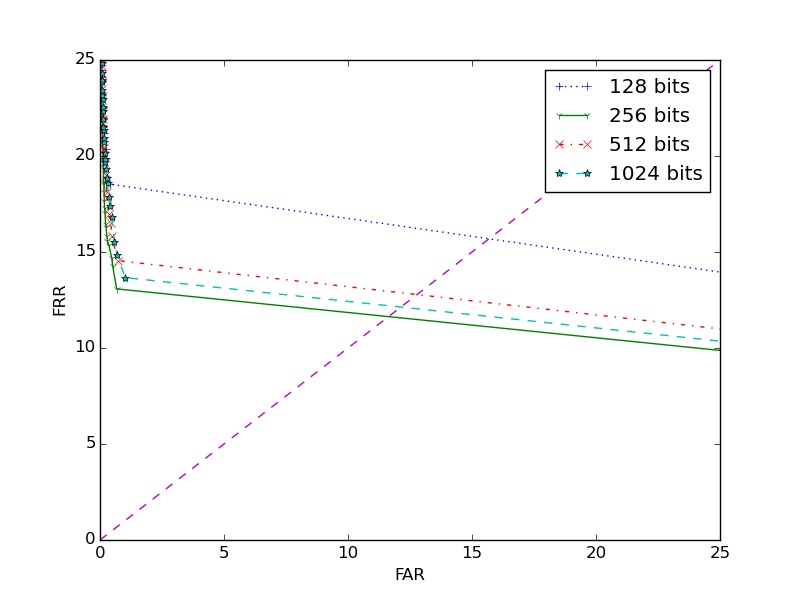}
	\end{center}
	\caption{ROC curves Extended Yale B (scenario 2) with varying bits of security.}
	\label{fig:yaleroc2}
\end{figure}

\begin{figure}
	\begin{center}
		\includegraphics[width=\linewidth]{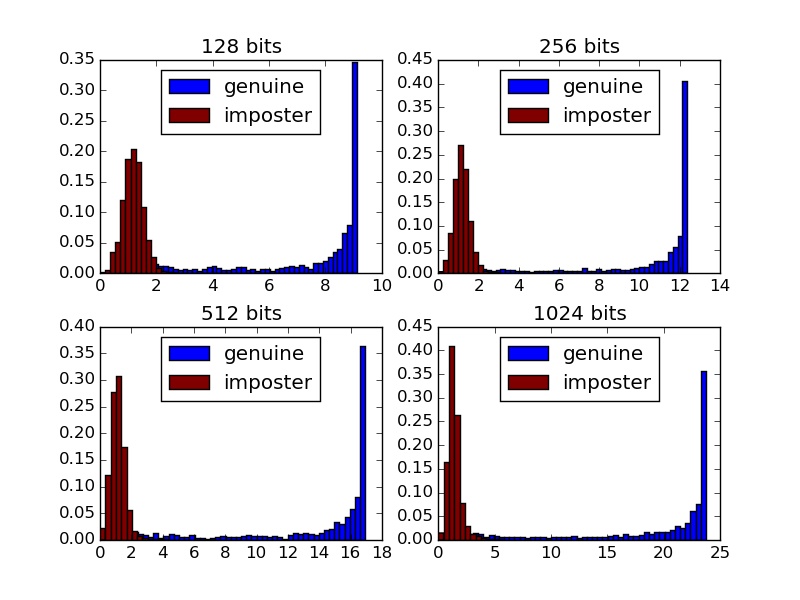}
	\end{center}
	\caption{Genuine and imposter distributions for Extended Yale B (scenario 1) with varying bits of security.}
	\label{fig:yalegenimp1}
\end{figure}

\begin{figure}
	\begin{center}
		\includegraphics[width=\linewidth]{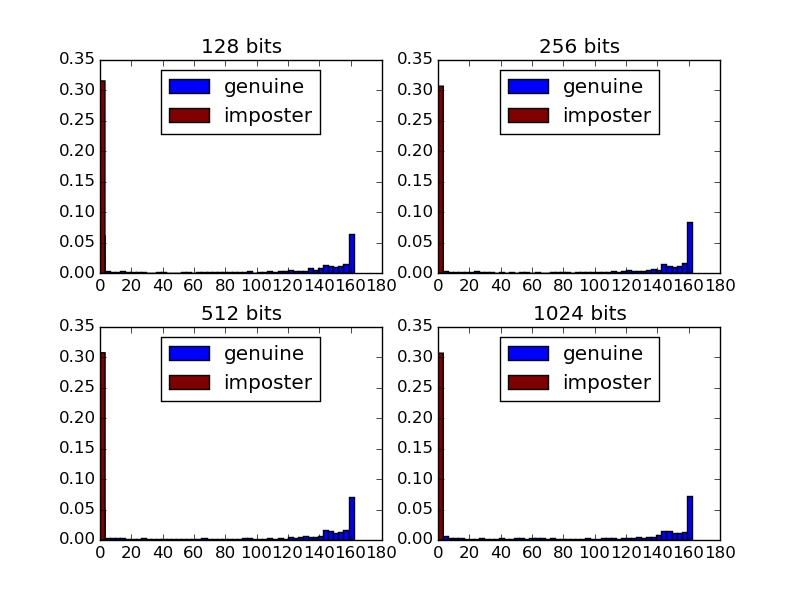}
	\end{center}
	\caption{Genuine and imposter distributions for Extended Yale B (scenario 2) with varying bits of security.}
	\label{fig:yalegenimp2}
\end{figure}

\begin{table*}[t]
	\begin{center}
		\begin{tabular}{|c|c|c|c|c|}
			\hline
			Bits of security & EER (scenario 1) & Accuracy (scenario 1) & EER (scenario 2) & Accuracy (scenario 2) \\ \hline \hline
			128 & 6.37\%                              & 85.20\%                                     & 15.65\%                              & 85.10\%                                     \\ 
			256 & 5.92\%                              & 86.77\%                                     & 11.63\%                              & 86.92\%                                     \\ 
			512 & 5.74\%                              & 84.37\%                                     & 12.78 \%                              & 84.22\%                                     \\ 
			1024 & 5.45\%                            & 84.86\%                                     & 12.13\%                              & 84.66\%                                     \\ \hline
		\end{tabular}
	\end{center}
	\caption{Identification accuracy and equal error rate on  Extended Yale B database with varying bits of security.}
	\label{tab:yale}
\end{table*}

\begin{table}[h]
	\begin{center}
		\begin{tabular}{|c|c|c|c|}
			\hline 
			Method & Bits of security & EER & Accuracy \\ \hline \hline
			Hybrid Approach \cite{feng2010hybrid} & 210                 & 6.81\%    & -        \\ 
			BDA \cite{feng2012binary} & 76               & -     & 93.30\%        \\ 
			DSE (scenario 1) & 256                 & \textbf{1.36}\%    & \textbf{95.93}\%        \\ 
			DSE (scenario 2 )& 256                 & \textbf{3.23}\%    & \textbf{95.68}\%        \\ \hline
		\end{tabular}
	\end{center}
	\caption{Comparison with state of the art methods on CMU-PIE dataset.}
	\label{tab:cmucomp}
\end{table}

\subsection{Databases}
The CMU PIE \cite{sim2002cmu} database consists of 41,368 images of 68 people under 13 different poses, 43 different illumination conditions, and with 4 different expressions. We use 5 poses (c27, c05, c29, c09 and c07) and all illumination variations for our experiments. 10 images are randomly chosen for training and the rest are used for testing.

The extended Yale Face Database B \cite{GeBeKr01} contains 2432 images of 38 subjects with frontal pose and under different illumination variations. We use the cropped version of the database for our experiments. Again, we use 10 randomly selected images for training and the rest for testing. 

\subsection{Evaluation metrics}
In order to evaluate the performance of our framework under both scenario 1 and 2, we use the following metrics. For verification performance, we use ROC curves, the equal error rate (EER), and the genuine and impostor score distributions to measure matching performance. The ROC curve is plotted between the false reject rate (FRR) and the false acceptance rate (FAR) at all possible values of score thresholds. Note that, in scenario 1, a real valued score is obtained thus, the value of the threshold at the EER is a valid operating point for the system. For scenario 2, the score is in fact a discrete value and thus, the EER value might not correspond to a threshold that can be set for the system. For both scenarios we read the EER value off the ROC curve in order to avoid assumptions regarding the slope between the points at which FAR and FRR are closest. For identification, we use the classification accuracy as a measure of performance.

\subsection{Experimental parameters}
We use the same training procedure for both databases. The original images are resized to $m\times m = 64\times 64$ pixels. All possible crops of size $n\times n = 56\times 56$ are used to augment the training set, yielding $2\times 9\times 9 = 162$ images for each training image. High entropy random binary codes are generated and assigned to each class. We vary the code length, and thus the bits of security of the system, from $k = 128 -1024$ for our experiments. Next, the network is trained to minimize the cross-entropy loss against these codes for $100$ epochs with a batch size of $200$. The SHA-256 hashes of the codes assigned to each class are stored as the templates.

During testing, all crops and their flipped versions are extracted from each sample ($162$ in all), and the outputs for each are calculated by feeding through the network. For scenario 1, the score is proportional to the negative of the mean of the euclidean distance between each crop's output and the stored code. In scenario 2, the output of the network for each crop is quantized into a binary vector using $0.5$ as the threshold. These binary vectors are now hashed and the hashes are compared with the stored hashes in the database. The score in this case is discrete, and given by the number of matching hashes for the sample. The genuine scores are calculated by matching against the true class where as the imposter scores are calculated by matching against all other enrolled classes apart from the true class.

\subsection{Results}
The EER and accuracy for both scenarios on CMU PIE are shown in Table \ref{tab:pie}. The ROC curves for scenario 1 and 2 at different bits of security are show in Figure \ref{fig:rocpie1} and Figure \ref{fig:rocpie2} respectively. Genuine and imposter score distributions for scenario 1 and 2 are shown in Figure \ref{fig:genimppie1} and Figure \ref{fig:genimppie2}. Similarly, the EER and accuracy results for Extended Yale B are shown in Table \ref{tab:yale}. ROC curves are shown in Figure \ref{fig:yaleroc1} and Figure \ref{fig:yaleroc2}, where as, genuine and imposter distributions are shown in Figure \ref{fig:yalegenimp1} and Figure \ref{fig:yalegenimp2}. It can be seen that the performance of the system is consistent even when the bits of security is significantly varied. We also compare our approach, Deep Secure Encoding (DSE), to previous results on CMU PIE at comparable bits of security in Table \ref{tab:cmucomp}. Note that a fair comparison of our system to previous work is scenario 1 since the score in this case is based on the distance from the template as in the case of \cite{feng2010hybrid, feng2012binary}. The distance based score od scenario 1 captures how far the sample is from the template even in cases in which the counting based score of scenario 2 would yield no matches. Even so, our performance in terms of matching in both scenarios is significantly better than previous approaches. Also, the security offered by the system in scenario 2 is higher and with fewer assumptions when compared to past approaches.


\begin{figure}
	\begin{center}
		\includegraphics[width=\linewidth]{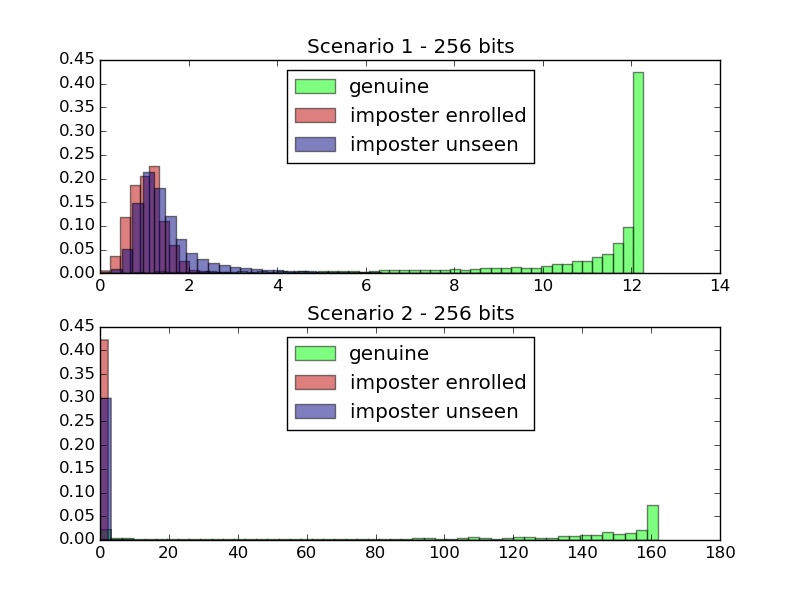}
	\end{center}
	\caption{Genuine and imposter distributions for enrolled and unseen imposters at 256 bits of security.}
	\label{fig:genimpunseen}
\end{figure}

\section{Security analysis}
We now discuss the security for both scenario 1 and 2. In scenario 1, the codes are stored without hashing and thus, given the codes and the network parameters, it is possible to retrieve some part of the original modality. Even though there would be loss of information propagating backwards through the network, this scenario is best used only when the network parameters can be protected. We consider this scenario to get a better idea of the distribution of the genuine and impostor scores in terms of the distance of the network output from the assigned codes, and also to make our results comparable to the continuous scores of other approaches.

In case of scenario 2, the SHA-256 hashes of the codes are stored as templates and thus, even if the attacker has both the templates as well as the network parameters, it is not possible to retrieve any part of the original data. The only possible attack on the template is that of brute forcing through all the possible codes. The complexity of this is of the order of $2^{k}$ where $k$ is the length of the binary codes. An important aspect of the security offered by scenario 2 is that, unlike previous works, we do not need to make any assumptions about the distribution of the data being hashed. When the hashed component of the system fits a low entropy distribution, the search space while brute forcing the templates is significantly reduced. In our case, the data being hashed is the set of random codes assigned to face classes. Since the codes are bit-wise randomly generated, they possess maximum entropy and the attacker would need to brute force a search space that is truly of the size of $2^{k}$. Our system performs well through a wide range of choices of $k$ making is highly secure against attacks.

Another concern is that of data that has not been seen by the model. The genuine and imposter score distributions thus far indicate that the scores are well separated for genuine and imposter users, but, our imposter set consists of all other enrolled classes apart from the genuine one. Thus, we do another experiment to test the performance of the system against imposters that have not been seen by the model. For this purpose, we train our model on 20 classes from CMU PIE and generate two sets of imposter scores, one using classes from within the 20 used for enrollment and another using all samples from the remaining 48 classes. We refer to these as imposter enrolled and imposter unseen and the score distributions for scenario 1 and 2 are shown in Figure \ref{fig:genimpunseen}. We can see that the imposter distribution for unseen imposters is almost identical to that of the enrolled imposters indicating the robustness of the system to unseen imposters.

Lastly, it is strightforward to see that cancelability can also be achieved by assigning a new set of random codes to user classes.

\section{Conclusion}
We presented a secure classification scheme that achieves state of the art matching performance, along with high security with no unrealistic assumptions. The work portrays the high learning capacity of deep networks and we address some of the issues related to limited data training in the application of deep learning to biometrics. Our future plans are to analyze other issues in the application of powerful deep learning architectures to the needs of the biometrics community. We also intend to test the system on more difficult face databases and other biometric modalities, working towards designing a multi-modal framework that is agnostic to the type of biometric modality it has to work with.


{\small
\bibliographystyle{ieee}
\bibliography{submission_example}
}

\end{document}